\documentclass{article} 
\usepackage{colm2024_conference}

\usepackage{microtype}
\usepackage{hyperref}
\usepackage{url}
\usepackage{booktabs}
\usepackage{graphicx}
\usepackage{amsmath}
\usepackage{amsthm}
\usepackage{booktabs}
\usepackage{algorithm}
\usepackage{algorithmic}
\usepackage{adjustbox}
\usepackage{multirow}

\usepackage{colortbl}
\usepackage{xcolor}
\usepackage{color}

\usepackage{enumitem}
\setlist{nolistsep}
\usepackage{xcolor}
\usepackage{diagbox}
\usepackage{makecell}
\usepackage{multirow}
\usepackage{booktabs}
\usepackage{geometry}
\usepackage{listings}
\usepackage{makecell}
\usepackage{multicol}
\usepackage{hyperref}

\newsavebox{\instruction}
\begin{lrbox}{\instruction}
\begin{minipage}{0.65\textwidth}
\begin{verbatim}
You can call the following tools:
{ "name": "search_engine",
 "description": "Search for information that will aid 
            in determining a response to the user.",
 "parameters": 
    {"type": "object",
    "properties": {"input": {"type": "string", 
                "description": "search keywords"}},
    "required": ["input"]} }
\end{verbatim}
\end{minipage}
\end{lrbox}

\title{Tool Calling: Enhancing Medication Consultation via \\ Retrieval-Augmented Large Language Models}

\author{Zhongzhen Huang\textsuperscript{1,2}, \quad Kui Xue\textsuperscript{2}, \quad Yongqi Fan\textsuperscript{3,2},  \quad Linjie Mu\textsuperscript{1}, \quad Ruoyu Liu\textsuperscript{2}, \\ \textbf{Tong Ruan\textsuperscript{3}, \quad Shaoting Zhang\textsuperscript{2,4}, \quad Xiaofan Zhang\textsuperscript{1,2*}} \\
\textsuperscript{1}Shanghai Jiao Tong University \quad
\textsuperscript{2}Shanghai AI Laboratory \\ \textsuperscript{3}East China University of Science and Technology 
\quad \textsuperscript{4}SenseTime Research \\
\texttt{\{huangzhongzhen, linjiemu, xiaofan.zhang\}@sjtu.edu.cn,} \\
\texttt{\{xuekui, liuruoyu, zhangshaoting\}@pjlab.org.cn}\\
\texttt{\{y21210043, ruantong\}@ecust.edu.cn}
}
\colmfinalcopy 
\begin{document}

\maketitle

\begin{abstract}
Large-scale language models (LLMs) have achieved remarkable success across various language tasks but suffer from hallucinations and temporal misalignment. To mitigate these shortcomings, Retrieval-augmented generation (RAG) has been utilized to provide external knowledge to facilitate the answer generation. However, applying such models to the medical domain faces several challenges due to the lack of domain-specific knowledge and the intricacy of real-world scenarios.
In this study, we explore LLMs with RAG framework for knowledge-intensive tasks in the medical field. To evaluate the capabilities of LLMs, we introduce MedicineQA, a multi-round dialogue benchmark that simulates the real-world medication consultation scenario and requires LLMs to answer with retrieved evidence from the medicine database. MedicineQA contains 300 multi-round question-answering pairs, each embedded within a detailed dialogue history, highlighting the challenge posed by this knowledge-intensive task to current LLMs. We further propose a new \textit{Distill-Retrieve-Read} framework instead of the previous \textit{Retrieve-then-Read}. Specifically, the distillation and retrieval process utilizes a tool calling mechanism to formulate search queries that emulate the keyword-based inquiries used by search engines. With experimental results, we show that our framework brings notable performance improvements and surpasses the previous counterparts in the evidence retrieval process in terms of evidence retrieval accuracy. This advancement sheds light on applying RAG to the medical domain. 
\end{abstract}

\section{Introduction}
Large language models (LLMs)~\citep{achiam2023gpt, touvron2023llama, team2023gemini} have revolutionized the field of natural language processing, showing remarkable impacts with the well-documented emergence of zero-shot capabilities in a variety of downstream tasks, like machine translation~\citep{zhang2023machine}, text generation~\citep{kojima2022large} and machine reading comprehension~\citep{samuel2023can}. Such impressive abilities stem from the ever-increasing number of parameters and large-scale training corpus.

Despite the massive knowledge, LLMs still struggle with considering issues of hallucination (i.e., prone to generate factually incorrect statements)~\citep{bang2023multitask, ji2023survey} and temporal misalignment (i.e., unable to capture the changing world)~\citep{kandpal2023large} in a set of tasks~\citep{yin2022survey, lewis2020retrieval}. Such knowledge-intensive tasks require access to a vast amount of knowledge beyond the training data.
Towards this issue, existing methods~\citep{li2023structure, jiang2023llmlingua, xu2023recomp, wang2023knowledgpt, cheng2024lift} incorporated external knowledge with LLMs by retrieval augmentation, dubbed as Retrieval Augmented Generation (RAG). In detail, LLMs retrieve the relevant information for the input query and utilize the retrieved evidence as additional context to generate the response. Such \textit{Retrieve-then-Read} framework cleverly combines flexible knowledge sources in a non-parameterized form for knowledge-intensive tasks and has become one of the hottest paradigms to alleviate the drawbacks in naive LLM generations. 



\begin{figure}
\centering
\scalebox{1}{
\includegraphics[width=\linewidth]{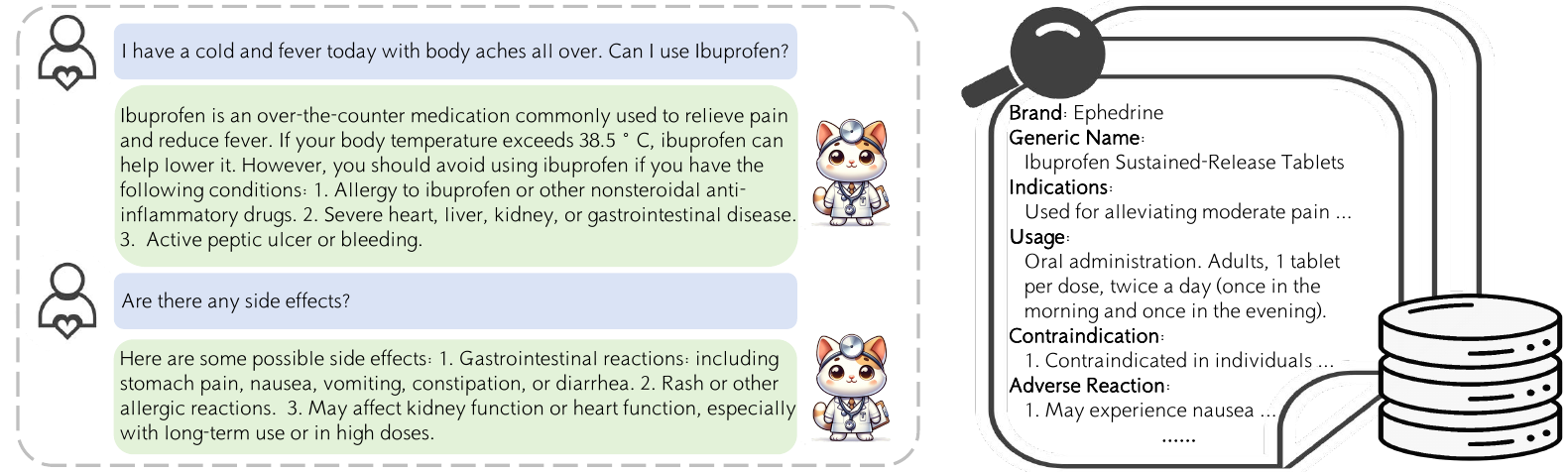}
}
\vspace{-5pt}
\caption{The medication consultation: a detailed discussion between healthcare professionals and users about prescribed medications, including their names, indications, usage, side effects, etc. Professionals utilize the knowledge in the medicine database to provide a more robust response.}
\label{fig:intro}
\vspace{-10pt}
\end{figure}

Beneath the advancements, we find a notable gap in applying LLMs to medical fields, especially for knowledge-intensive tasks, like medication consultation. As shown in Figure~\ref{fig:intro}, medication consultation aims at providing real-time accessibility for medication-related inquiries and enhancing medication safety through searching from the database, requiring depth in domain-specific areas. In real-world scenarios, the dialogs are usually ambiguous and verbose, e.g., users tend to use layman's terms instead of standard terms and provide much more information than what might be medically relevant.
We ask: \textit{Is the LLM with vanilla RAG enough for the medication consultation?} 

In this work, we introduce a new benchmark, MedicineQA, to evaluate the proficiency of LLMs in medication consultation scenarios. We recruited a panel of 5 board-certified physicians to create the benchmark as follows: sourcing and rephrasing questions from an online medical consultation website; simulating multiple rounds of dialogue scenarios via GPT-4~\citep{achiam2023gpt}. Our research reveals that vanilla RAG methods suffer from serious challenges in retrieving relevant information with intricate dialogue history. 

Based on PULSE~\cite {pulse2023}, we propose RagPULSE via the search engine tool. Instead of the \textit{Retrieve-then-Read} framework adopted by previous retrieval-augmented work, RagPULSE utilizes a novel \textit{Distill-Retrieve-Read} framework to access the external knowledge. Specifically, RagPULSE processes a medication inquiry by summarizing the dialogue history to keywords for searching API calls and integrating the retrieved evidence from the medicine database to formulate a comprehensive response.



Our main contributions can be summarized as follows:
\begin{itemize}
    \item We present MedicineQA, a new benchmark derived from real-world medication consultation, aimed at evaluating LLMs’ ability in the medical domain.
    \item We propose a pioneering retrieval augmentation framework, \textit{Distill-Retrieve-Read}, via the ``tool calling'' mechanism.
    \item Incorporated with the framework, our proposed RagPUSLE outperforms all publicly available models in performance and is competitive with state-of-the-art commercial products with a smaller parameter size.
\end{itemize}

\section{Related Work}
\textbf{Large Language Model in Medical Domain.} 
The impressive abilities of large language models (LLMs) across various applications have catalyzed extensive investigation into employing them in healthcare and medical domains. This surge in attention is documented through a growing body of research~\citep{thirunavukarasu2023large, clusmann2023future}. Some recent works have studied to augment LMMs with real-world data.  ChatDoctor~\citep{yunxiang2023chatdoctor}, trained by fine-tuning LLaMA~\citep{touvron2023llama} on a large dataset of patient-doctor dialogues, achieves high accuracy and reliability in medical scenarios with an external information retrieval module. From the other line, some adopt the synthetic data for fine-tuning. \cite {zhang2023huatuogpt} utilized real-world data from medical professionals alongside distilled data from ChatGPT to fine-tune the model. To enhance the capability in the multi-round conversation, BianQue~\citep{chen2023bianque} trained the model on a self-constructed dataset containing multi-round inquiries and health suggestions. Despite the remarkable performance, there is still a gap in applying LLMs in real-world scenarios due to the lack of domain-specific knowledge. To further evaluate the proficiency of LLMs in medical domains, we introduce MedicineQA, a benchmark derived from real-world medication consultation scenarios.


\textbf{Retrieval-Augmented Generation.} 
LLMs require external knowledge to alleviate the factuality drawbacks.
Retrieval-augmented generation (RAG) has been regarded as an effective solution to mitigate the aforementioned hallucinations and temporal misalignment issues inherent in large language models, especially for knowledge-intensive tasks.
Generally, studies of RAG can be categorized into three types~\citep{gao2023retrieval}, namely Naive RAG, Advanced RAG, and Modular RAG. Naive RAG means a straightforward \textit{Retrieve-then-Read} framework~\citep{lewis2020retrieval, karpukhin-etal-2020-dense, izacard2022few}. To enhance retrieval quality, the Advanced RAG builds upon the foundation of Naive RAG by incorporating pre-retrieval~\citep{li2023structure} and post-retrieval~\citep{jiang2023llmlingua, xu2023recomp} strategies. Modular RAG improves the overall performance by decomposing the \textit{Retrieve-then-Read} framework into fine-grained modules with distinct functionalities, such as a search module\citep{wang2023knowledgpt}, memory module\citep{cheng2024lift}.


\section{Method}
In Section~(\ref{subsec:data}), we propose MedicinceQA, a novel benchmark to evaluate LLMs' capabilities toward knowledge-intensive tasks in medical fields. We curate the benchmark from various real-world medication consultation scenarios and unified them into multi-round dialogue. Then, we present RagPULSE in Section~(\ref{subsec:rag}), a dedicated pipeline that adopts \textit{Distill-Retrieve-Read} framework for multi-round medication consultation. The fundamental operations of RagPULSE comprise three main steps: (1) the LLM calls the search engine tool and distills the dialogue history into a new query to gather evidence from the external medicine database; (2) the generated search query is executed to retrieve related evidence following a hierarchical form; (3) the retrieved evidence is provided to the LLM, and the LLM respond the user's question by the retrieved evidence.

\subsection{Benchmark Creation}
\label{subsec:data}
Existing benchmarks for evaluating the capabilities of LLMs in medical fields primarily focus on widely known or widely available tasks given a specific context (e.g., Automatic Structuring of medical reports and Named Entity Recognition). However, these benchmarks are insufficient for assessing LLMs' proficiency in knowledge-intensive tasks. Therefore, we introduce MedicineQA, a novel benchmark designed for evaluating LLMs within the context of medication consultation.

\textbf{Data Collection.} In an effort to align the benchmark with real-world scenarios, we crawled data from websites for medical consultation, which comprise numerous online consultation records between users and medical experts. Each record contains multiple rounds of dialogue, we categorized each record into three categories: 1) Diagnostic Process, where the expert diagnoses based on symptoms provided by the user; 2) Medication Consultation, where the expert addresses queries regarding medications for certain conditions; 3) Other, which includes the patient's medical history and some trivial communication. In total, we amassed 1,028,090 records comprising 6.24M pairs.

\textbf{Data Refinement.} Given the crawled data, we first conducted an initial statistical analysis and identified the 200 most commonly mentioned medicines as the scope for further processing.
To ensure the correctness, we recruited a panel of 5 board-certified physicians to curate the content.
The physicians filtered out irrelevant dialogues of each selected record and summarized it into one question about a specific medicine. 
For each summarized question, we utilized GPT-4~\cite {achiam2023gpt} to expand them into multi-round dialogue. Subsequently, physicians manually revised the dialogues to ensure a logical progression of questions, with each answer building on the information provided in the preceding dialogues and without repeating information. This process yielded 300 multi-round dialogue questions focused on medication consultation. 

\textbf{Medicine Database.} 
To provide precise and structured information, we introduce an entity-oriented medicine database with 42764 medicines, where each medicine is represented in three forms: brand name, generic name, and detailed attributes like usage, contraindications, adverse reactions, etc. Formally, for each medicine $M_i$ in our database $D$, we first concatenated its generic name with each attribute $a_j$ to obtain the entity-attribute items $E_{ij}$, respectively. Then, each item is embedded into vectors and stored in a tree form according to the entity, i.e., the information of the medicine $M_i$ is stored in the form of $E_i=\{E_{i1}, E_{i2}, E_{i3}, \dots\}$, accompanied by its corresponding keys $K^n_i$ and $\{K_{i1}^{a}, K_{i2}^{a}, K_{i3}^{a}, \dots\}$. In our database $D$, $E_i$ and $E_{ij}$ can be obtained via $D[K^n_i]$ and $D[K_{ij}^{a}]$, respectively.


\begin{figure}[th]
\centering
\scalebox{1}{
\includegraphics[width=\linewidth]{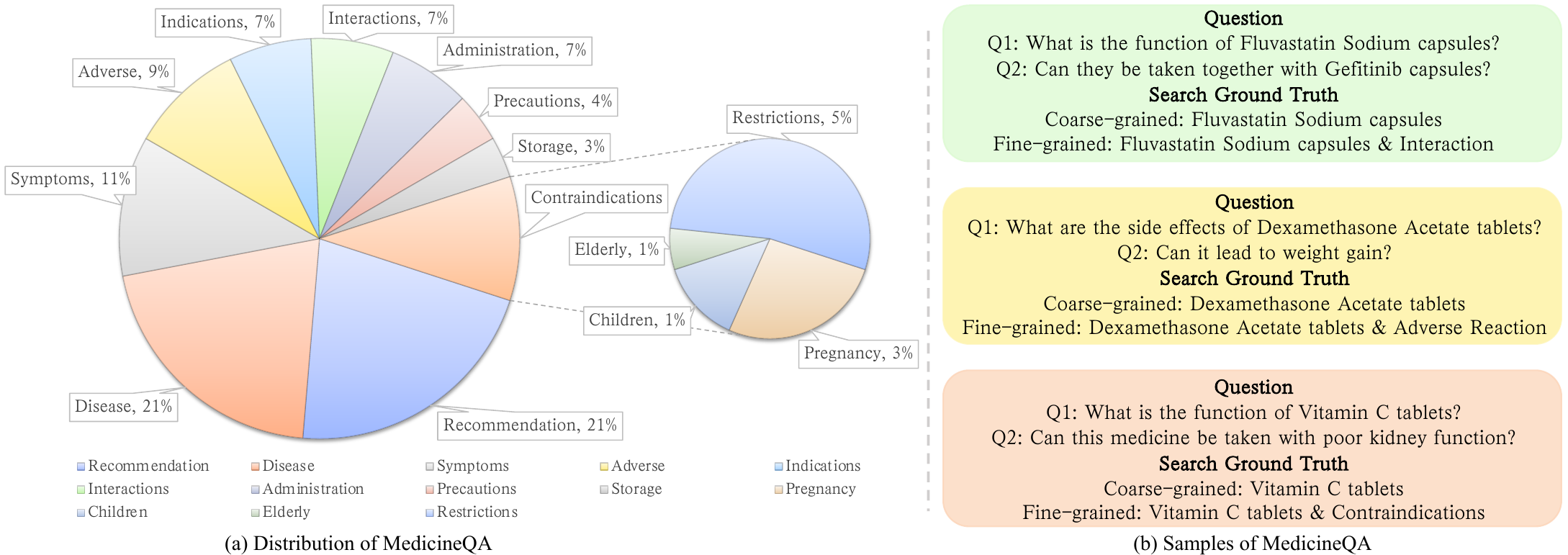}
}
\caption{(a) The distribution of our proposed MedicineQA. MedicineQA involves ten specific scenarios of the medication consultation. The distribution of the benchmark is similar to that of the real scenario. (b) Samples of the benchmark: \textcolor{green!50}{Interaction}, \textcolor{yellow!95}{Adverse reactions}, and \textcolor{orange!60}{Contraindications}. Our benchmark is available in both English and Chinese.}
\label{fig:data}
\end{figure}

\textbf{Annotation.}
In our benchmark, each question is associated with the corresponding medicine descriptions extracted from the medicine database, to serve as the retrieved evidence. To evaluate the retrieval process, we further labeled two types of retrieval ground truths: one is the document-level for coarse-grained evaluation $K_c$, and the other is the specific sections in the relevant documents for fine-grained attribute-level assessment $K_f$. One sample of our MedicineQA can be formulated as $\mathbf{S} = <{H, Q_{T+1}, K_c, K_f}>$, where $H=\{(Q_i, A_i)\}, i = 1, 2, \dots, T$ is the dialogue history, $(Q_i, A_i)$ denotes a round of conversation between the user and the agent, and $T$ is the number of dialogue rounds. $Q_{T+1}$ represents a question about one specific medicine. $K_c, K_f$ are the coarse-grained and fine-grained ground truth for evaluating the retrieval process, respectively. In detail, $K_c$ is the $K^n_i$ in $D$, and $K_f$ is a subset of $\{K_{i1}^{a}, K_{i2}^{a}, K_{i3}^{a}, \dots\}$. We display the relative distribution of our proposed benchmark and present samples of the created data in Figure~\ref{fig:data}.

\subsection{RagPULSE}
\label{subsec:rag}

\begin{figure}[t]
\centering
\scalebox{1}{
\includegraphics[width=\linewidth]{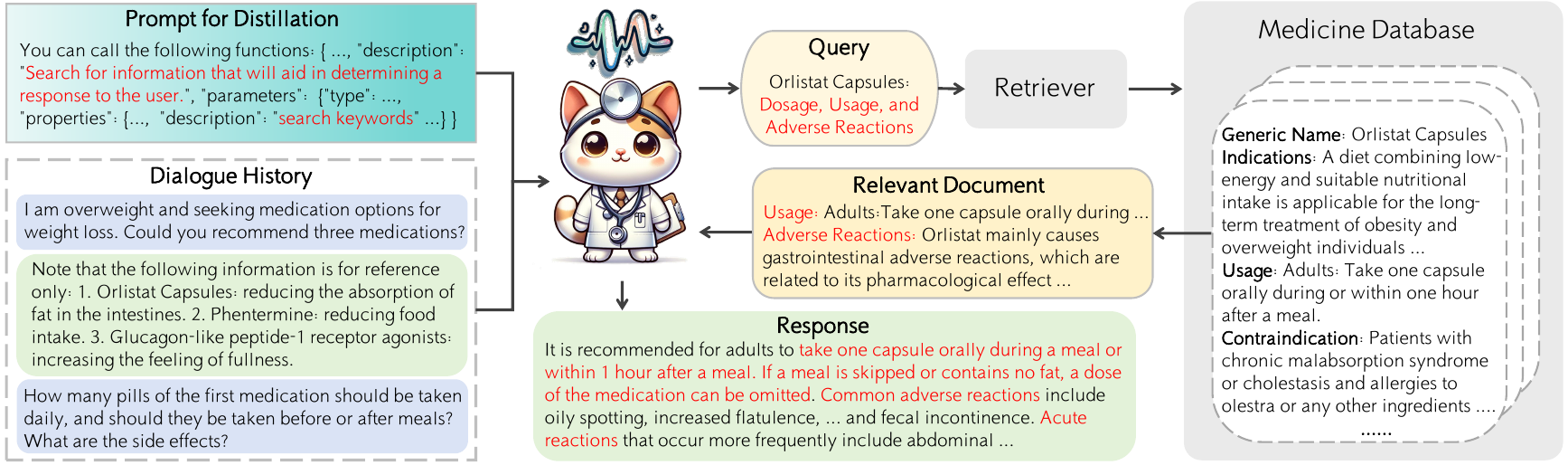}
}
\caption{The overall workflow of our RagPULSE in the medication consultation scenario, consists of three steps: (1) Distilling the key information and forming the searching query from the dialogue history; (2) Retrieving the corresponding medicine evidence from the medicine database; (3) Generating the response according to the retrieved evidence.}
\label{fig:Network}
\end{figure}


We choose PULSE~\citep{pulse2023} as the LLM, which demonstrates impressive performance in the medical field, and augment it with the \textit{Distill-Retrieve-Read} framework. As shown in Figure~\ref{fig:Network}, the process can be formulated into three steps. The LLM is first tasked to call the search engine tool and summarize the search query supported by the combination $[H, Q_{T+1}]$. Subsequently, the search engine retrieves relevant keys $\hat{K}$ from the medicine database $D$ and obtains the evidence $\hat{E}$ from the medicine database $D$. Finally, the LLM generates the answer $A_{T+1}$ according to $[H, Q_{T+1}, \hat{E}]$. 

\textbf{Tool Calling.} A simple but robust retrieval query is vital to clarify the search need from the context and eliminate irrelevant information in the external knowledge base. Recent studies either directly adopt the query from the dataset~\citep{liu2024optimizing} or rewrite it by the black-box generation~\citep{ma2023query}. However, there is inevitably a gap between the query and the evidence that needs to be obtained, especially for such a task with a long context. Only relying on the original capability of the LLM and human-written prompt lines makes it difficult to summarize correct inquiries from the intricate context while preserving key information. Inspired by the \textit{program of thought (PoT)}~\citep{chen2022program}, where the LLM generates Python code for retrieving, we integrate ``tool calling'' with the LLM. This approach prompts the LLM to generate search keywords for search tools, mimicking the use of search engines. With the above paradigm, the LLM is able to call the search tool and generate the retrieval query according to the current dialogue.


\textbf{Synthetic Dataset.} To endow the LLM with the distillation ability, we construct a synthetic dataset for the dialogue distilling task following previous works~\citep{ma2023query, hsieh2023distilling, ho2022large}. First, we collect a large-scale question set (including but not limited to dialogue questions and search engine questions) from several websites (e.g., Google and Baidu).  Then, the selected questions are distilled and summarized as pseudo labels by prompting GPT-4~\citep{achiam2023gpt} to utilize function call. After fine-tuning, the LLM shows remarkable performance in distilling the context into simple inquiries containing key information. The samples of synthetic data and the instructions for ``tool calling'' are shown in Table~\ref{tab:ins}.


\begin{table}[]
\label{tab:ins}
\adjustbox{max width=\textwidth}{%
\begin{tabular}{c|c}
\toprule
The Template of instructions for Tool Calling & Samples of Synthetic Data\\
\cmidrule(lr){1-1}\cmidrule(lr){2-2}
\multirow{9}{*}{\usebox{\instruction}} & \multirow{3}{*}{\makecell[c]{\textbf{Input}: 2017 college entrance examination ticket, fully opened, \\ how much longer? How wide is it? \\ \textbf{Output}: search\_engine(2017 College entrance examination ticket size.)}} \\
\\
\\
\cmidrule(lr){2-2}
& \multirow{3}{*}{\makecell[c]{\textbf{Input}: How much does it cost for high school students to study in Japan? \\ \textbf{Output}: search\_engine(The cost of studying in Japan high school.)}} \\
\\
\\
\cmidrule(lr){2-2}
& \multirow{3}{*}{\makecell[c]{\textbf{Input}: When is there a typhoon in Guangzhou? \\ \textbf{Output}: search\_engine(Guangzhou Typhoon Forecast.)}} \\
\\
\\
\bottomrule
\end{tabular}
} 
\caption{The instructions and samples of the synthetic dataset for fine-tuning the LLM.}
\label{table:prompt}
\end{table}

\section{Experiments}
In this section, we measure the performance of RagPULSE on MedicineQA and compare it to existing LLMs and commercial products (\ref{sec:Results}). We ablate the \textit{Distill-Retrieve-Read} on the MedicineQA dataset, showing their importance (\ref{sec:ab}). Finally, we present some cases to investigate the hallucinations of LLMs towards medication consultation.
\subsection{Experimental Settings}
\textbf{Implementation Details} 
We develop RagPULSE with \textit{Distill-Retrieve-Read} framework in Pytorch~\citep{paszke2019pytorch} and fine-tune it by the proposed synthetic dataset. It is worth noting that a single
machine with eight NVIDIA A100 GPUs proved sufficient for the memory requirements of PULSE~\citep{pulse2023}. Our training framework integrates tensor parallelism~\citep{wang2022tesseract} and ZeRO-powered data parallelism~\citep{rajbhandari2020zero}. 
To further accelerate training without sacrificing accuracy, we implement mixed-precision training, where we execute forward and backward computations in BFloat16 and conduct optimizer updating in Float32. For the compared models, we adopt the pre-trained weights and settings provided on the official website.

\textbf{Baselines.} Given the variety of current LLMs and the fact that MedicineQA is the medical domain, we choose open-sourced models and commercial products with notable performance in the medical domain to fully explore the current proficiency of LLMs in medication consultation scenarios. For a fair comparison, we utilize models that the results can be reproduced as follows: DoctorGLM~\citep{xiong2023doctorglm}, ChatGLM3~\citep{du2022glm}, BianQue2~\citep{chen2023bianque}, MING~\citep{MING}, QWen2~\citep{qwen}, Baichuan2~\citep{baichuan2023baichuan2} and ChatGPT3.5~\footnote{\href{}{https://chat.openai.com}}

    

\textbf{Metrics.} To evaluate the accuracy of the evidence retrieval stage, we employ the Hit Rate (HR@num), which represents the proportion of instances where the retrieval candidates contain the corresponding knowledge, with ``num'' indicating the number of candidates to be retrieved. We respectively calculate the hit rate of coarse-grained and fine-grained retrieval through the retrieved database key and the search ground truth. Given the answer of the medication consultation is in the form of free text, which is a challenge for evaluating the correctness, we utilize the Elo rating system~\citep{elo1967proposed,chiang2023vicuna,dettmers2023qlora} to gauge the performance of LLMs on MedicineQA. It adjusts a player's rating based on the outcome of their games, taking into account the expected score versus the actual score. In our settings, each model is one competitor, and the powerful GPT-4~\citep{achiam2023gpt} serves as the referee to determine which model performs better. More details can be seen in the Appendix. 

\begin{table*}[tbp]
    \centering
    \renewcommand\arraystretch{1.1}
    \resizebox{\linewidth}{!}
    {
    \begin{tabular}{@{}l c c |ccc|ccc|c c}
    \toprule
         \multirow{2}{*}{Model Name} & \multirow{2}{*}{\makecell[c]{Param.\\Size}} &\multirow{2}{*}{\makecell[c]{Ins. follow\\ rate (\%)}}  & \multicolumn{3}{c|}{Retrieved Doc. (\%)}  & \multicolumn{3}{c|}{Retrieved Attr. (\%)} & \multicolumn{2}{c}{Generation} \\
         \cmidrule{4-11}
         & &  & HR@1 & HR@5 & HR@10 & HR@1 & HR@5 & HR@10 & Elo Rating & Elo Rank\\ \midrule
         BianQue2& 6B &3.33 &7.33 & 9.00 & 10.00&1.67 & 2.00 & 2.00 &883 &10\\
         DoctorGLM& 6B &47.00 &12.67 & 15.00 & 16.00 &2.33 & 2.67 & 3.00 &920 &8\\ 
         ChatGLM3& 6B &92.33 &27.33 & 32.00 & 34.00 &8.00 & 9.33 & 9.67 &999 &7\\
         MING& 7B &8.00 &20.00 & 28.33 & 30.67 & 5.67 & 7.67 & 8.00&1017 &6\\
         BenTsao& 7B &16.67 &33.33 & 45.33 & 48.00 &12.67 & 17.33 & 18.33 &913 &9\\
         Baichuan2& 14B &98.33 &52.67 & 66.67 & 71.33 &26.67 & 35.33 & 38.00& 1045&4\\
         QWen2& 14B &100.00 &57.67 & 68.33 & 76.67&25.33 & 28.33 & 30.33& 1018&5\\
         ChatGPT3.5& - &100.00 &63.67 & 72.33 & 78.67&27.00 & 31.33 & 32.67& 1072&2\\ 
         \rowcolor{blue!8} RagPULSE& 7B &100.00 &63.67 & 73.00 & 78.33 &\textbf{28.33} & \textbf{32.00} & \textbf{33.33} &1058 &3\\
         \rowcolor{blue!8} RagPULSE& 20B &100.00 &\textbf{65.67} & \textbf{75.33} & \textbf{78.33} &27.33 & 31.67 & 32.33&\textbf{1074} &\textbf{1}\\
    \bottomrule     
    \end{tabular}
    }
    \caption{Evaluation on MedicineQA. Our study employs the PULSE model with varying parameter sizes, augmented by the \textit{Distill-Retrieve-Read} framework. We compare them with other LLMs and commercial products. 
    ``Retrieved Doc.'' refers to the process of only searching the generic name of the medicine (coarse-grained), while ``Retrieved Attr.'' denotes calculating the results via the combination of the generic name and the specific attribute (fine-grained).}
    \label{tab:main}
\end{table*}

\subsection{Results}
\label{sec:Results}
Here we thoroughly evaluate models using the MedicineQA benchmark. To assess the performance of evidence retrieval, we prompt those baseline models to formulate search queries by summarizing preceding dialogues and then calculate their accuracy in retrieving relevant evidence. Due to the limitations of some baseline models in retrieving evidence from the medicine database, we immediately adopt the attached corresponding medicine information as the context to guide the generation of the final responses. It is worth noting that our RagPULSE leverages the retrieved evidence to generate the answer. Experimental results are reported in Table~\ref{tab:main}. 

From Table~\ref{tab:main}, we can see that some open-sourced models with smaller model sizes suffer from following the instructions for summarizing key information in specific format from complex dialogue histories, highlighting the inherent difficulties in medication consultation tasks. Finetuned on the synthetic dataset, our RagPUSLE (7B) presents a surprising performance in the instruction following rate. This outcome validates the effectiveness of adopting the code form of ``tool calling,'' underscoring the potential benefits of integrating programming paradigms into LLMs to bolster their understanding and execution of complex tasks.
As shown in Table~\ref{tab:main}, the \textit{Distill-Retrieve-Read} framework brings performance gains for the evidence retrieval process. Incorporated with the ability to distill dialogue history, RagPULSE is capable of summarizing the retrieval query. Compared with models whose number of parameters is less than 7 billion, RagPUSLE (7B) demonstrates a notable performance enhancement in the context of retrieval accuracy, achieving at least a 30\% improvement in document retrieval and a 15\% increase in attribute retrieval according to HR@1 metrics. This shows that some of the current open-sourced LLMs still struggle with distilling key information from the long context for searching relevant evidence. Regarding the models with more parameters, RagPUSLE (7B) still maintains a substantial lead, as evidenced by a 5\% improvement in HR@1. Surprisingly, RagPUSLE (7B) surpasses all models in attribute retrieval and RagPUSLE (20B) performs better than ChatGPT (65.67 vs. 63.67 in document retrieval ). These results indicate that using ``tool calling'' to distill context benefits the query generation. Moreover, we can see that RagPULSE outperforms all competing models and products in terms of responding to medication consultation even with the retrieved evidence. Depending on the remarkable capabilities of PULSE in the medical field, RagPULSE achieves a higher score than other open-sourced models.

Additionally, RagPULSE distinguishes itself across all metrics in the domain of medication consultation responses, even when utilizing retrieved evidence. Attributable to the specialized proficiency of PULSE in medical contexts, RagPULSE attains higher performance metrics than other publicly available models. This evidence highlights the superior capability of the \textit{Distill-Retrieve-Read} framework in processing and responding to complex medical inquiries, reinforcing its value in enhancing the accuracy and relevance of evidence retrieval in this specialized field.

\begin{table*}[tbp]
    \centering
    \renewcommand\arraystretch{1.0}
    \resizebox{\linewidth}{!}
    {
    \begin{tabular}{@{}l c|cccc|ccc c}
    \toprule
         \multirow{2}{*}{Model Name} & \multirow{2}{*}{\makecell[c]{Param.\\Size}} & \multicolumn{4}{c|}{Retrieved Doc. (\%)}  & \multicolumn{4}{c}{Retrieved Attr. (\%)} \\
         \cmidrule{3-10}
         &  & HR@1 & HR@5 & HR@10 & HR@50 & HR@1 & HR@5 & HR@10 & HR@50\\ \midrule
         History& - &18.33 & 27.00 & 31.00& 40.33 &5.33 & 6.67 & 7.67 & 9.00  \\
         Last Question &- & 28.33 & 35.00 & 37.67& 40.00 &12.33 & 15.67 & 16.33 & 17.67 \\
         \rowcolor{red!8}PULSE& 7B &53.00 & 62.67 & 66.00&70.33
  &18.00 & 21.00 & 22.00 & 23.33\\ 
         \rowcolor{red!8}RagPULSE$^{\dag}$& 7B &58.67 & 69.67 & 75.67& 78.67
 &19.67 & 22.67 & 23.67 & 25.00\\ 
         \rowcolor{red!8}RagPULSE& 7B &63.67 & 73.00 & 78.33 &82.00 &28.33 & 32.00 & 33.33 & 35.00\\
         \rowcolor{yellow!8}PULSE& 20B &56.33 & 66.33 & 69.67& 74.00 &22.00 & 26.33 & 26.67&28.00\\
         \rowcolor{yellow!8}RagPULSE$^{\dag}$& 20B & 60.33 & 70.67 & 75.00&81.00
  &29.33 & 34.00 & 34.67 & 38.67\\
         \rowcolor{yellow!8}RagPULSE& 20B &65.67 & 75.33 & 78.33&82.33
  &27.33 & 31.67 & 32.33&35.33
\\
    \bottomrule     
    \end{tabular}
    }
    \caption{Ablation of the \textit{Distill-Retrieve-Read} framework. The ``History'' setting implements the retrieval process by using dialogue history as the query and the ``Last Question'' setting conducts searching via the last question. We also prompt RagPULSE by the instruction used for baseline models, which are denoted as $^{\dag}$. }
    \label{tab:ab}
\end{table*}

\subsection{Ablation Studies}
\label{sec:ab}
To fully investigate the contribution of our proposed \textit{Distill-Retrieve-Read} framework, we conduct a quantitative analysis and report performances on MedicineQA when toggling the distillation part. The first two rows of Table~\ref{tab:ab} underscore the importance of distilling key information from dialogue history, which otherwise includes extraneous details detrimental to effective evidence retrieval. In addition, relying solely on the most recent query for information search proves inadequate due to the critical context embedded within the dialogue. Notably, RagPULSE (7B) exhibits more pronounced improvements, which outperforms PULSE (7B) with a notable 10\% improvement. 

Furthermore, as in the previous experiments, we also prompt our models to summarize the keywords without calling the tool. Compared with the PULSE without fine-tuning, RagPULSE$^{\dag}$ are observed to have significant performance gains in the two retrieval results. The results validate the effectiveness of our proposed synthetic dataset for summarizing the history and confirm that fine-tuning models on our synthetic dataset can endow models with distillation abilities.
\begin{figure}[t]
\centering
\scalebox{1}{
\includegraphics[width=\linewidth]{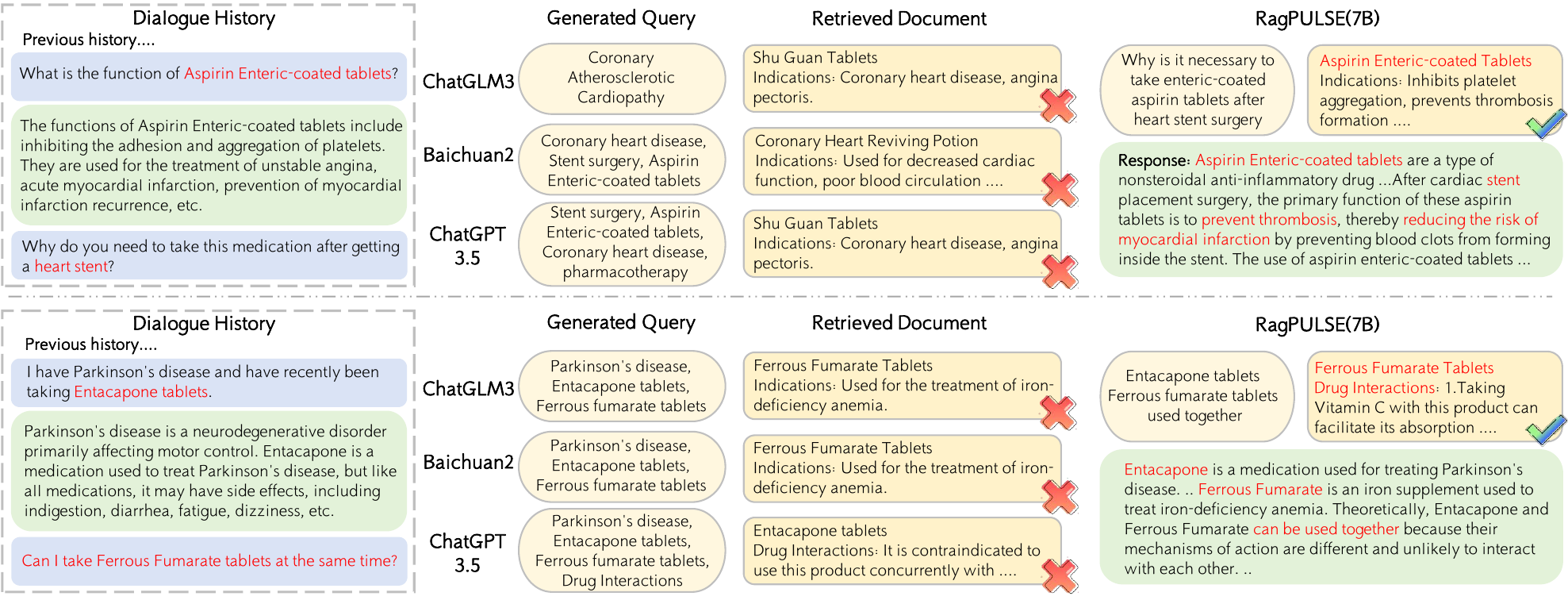}
}
\caption{Case studies of LLMs' retrieval process and generated responses. LLMs first summarize the dialogue history and then generate search queries. The responses are formulated via the retrieved document. Key information is marked by \textcolor{red}{red text}. }
\label{fig:case}
\end{figure}
\subsection{Case Study}
To intuitively show how the \textit{Distill-Retrieve-Read} framework makes a difference in the evidence retrieval process, we present examples (i.e., ChatGLM3, Baichuan2, ChatGPT3.5, and RagPULSE-7B) in Figure~\ref{fig:case} to compare the generated searching queries and the retrieved evidence. As can be seen in the upper part, in scenarios involving lengthy history, extraneous information often leads to the generation of redundant and ineffective search queries. It is evident that, despite LLMs' ability to generate queries encapsulating all necessary information, the complexity of such queries frequently results in retrieval failures. In the lower part, although the query contains the corresponding medicine, the LLMs fail to understand the question, resulting in the omission of crucial keywords. Additionally, we can observe that ChatGPT3.5 still fails despite generating the correct keywords since the query does not contain key information about the question. These examples clearly indicate the state of current LLMs in the medication scenarios. With supplemented knowledge, RagPULSE shows hopeful performance in generating responses for medication consultation. 

\section{Conclusion}
In this paper, we introduce MedicineQA, a new benchmark derived from real-world medication consultations, which aims at evaluating the capabilities of LLMs towards knowledge-intensive tasks in the medical domain. Our study shows that the LLM with vanilla RAG is not enough for the medication consultation. To address this, we propose RagPULSE with a novel framework, \textit{Distill-Retrieve-Read}, which revolutionizes the conventional \textit{Retrieve-then-Read} through the innovative use of the ``tool calling'' mechanism. Extensive experiments demonstrate that our model gains superior performance compared to existing models in two evidence retrieval processes. Furthermore, integrated with an entity-oriented medicine database, our RagPULSE presents impressive results in responding to inquiries in medication consultation. We hope our work can motivate further innovation in applying LLMs in the medical domain.

\newpage
\bibliography{colm2024_conference}
\bibliographystyle{colm2024_conference}

\appendix
\section{Appendix}
\subsection{Details of Elo}
The Elo rating system, devised by Arpad Elo, is a methodical framework used to calculate the relative skill levels of players in competitor-versus-competitor games. Initially conceived for chess, the Elo system has found widespread application across various sports and games to gauge individual or team performance. The fundamental principle of the Elo system is to assign a numerical rating to each player, which adjusts based on match outcomes against other rated players. The adjustment in ratings is predicated on the difference between the actual and expected match outcomes, allowing for a dynamic representation of a player's skill level over time.

The core of the Elo rating system is encapsulated by the formula used to update player ratings post-match. The expected score for a player, $E_A$, against an opponent, is calculated as:
$$
E_A=\frac{1}{1+10^{\left(R_B-R_A\right) / 400}}
$$
where $R_A$ and $R_B$ are the current ratings of the player and the opponent, respectively. Following the completion of a match, the actual score $\left(S_A\right)-1$ for a win, 0.5 for a draw, and 0 for a loss -is compared against the expected score to update the player's rating:
$$
R_A^{\prime}=R_A+K\left(S_A-E_A\right)
$$
In this formula, $R_A^{\prime}$ represents the new rating of the player, and $K$ is a factor that determines the maximum possible adjustment per game. This factor can vary depending on the level of competition and the governing body's regulations, allowing for flexibility in the sensitivity of rating adjustments to match outcomes. The Elo system's adaptability and simplicity have contributed to its enduring popularity and applicability across different competitive disciplines.
\end{document}